\PassOptionsToPackage{unicode}{hyperref}
\PassOptionsToPackage{hyphens}{url}
\documentclass[
]{article}
\usepackage[a4paper,left=0.7in,right=0.7in,top=0.8in,bottom=0.8in]{geometry}
\usepackage{xcolor}
\usepackage{amsmath,amssymb}
\usepackage{iftex}
\ifPDFTeX
  \usepackage[T1]{fontenc}
  \usepackage[utf8]{inputenc}
  \usepackage{textcomp} % provide euro and other symbols
\else % if luatex or xetex
  \usepackage{unicode-math} % this also loads fontspec
  \defaultfontfeatures{Scale=MatchLowercase}
  \defaultfontfeatures[\rmfamily]{Ligatures=TeX,Scale=1}
\fi
\usepackage{lmodern}
\ifPDFTeX\else
\fi
\IfFileExists{upquote.sty}{\usepackage{upquote}}{}
\IfFileExists{microtype.sty}{% use microtype if available
  \usepackage[]{microtype}
  \UseMicrotypeSet[protrusion]{basicmath} % disable protrusion for tt fonts
}{}
\makeatletter
\@ifundefined{KOMAClassName}{% if non-KOMA class
  \IfFileExists{parskip.sty}{%
    \usepackage{parskip}
  }{% else
    \setlength{\parindent}{0pt}
    \setlength{\parskip}{6pt plus 2pt minus 1pt}}
}{% if KOMA class
  \KOMAoptions{parskip=half}}
\makeatother
\usepackage{longtable,booktabs,array}
\usepackage{calc} % for calculating minipage widths
\usepackage{etoolbox}
\makeatletter
\patchcmd\longtable{\par}{\if@noskipsec\mbox{}\fi\par}{}{}
\makeatother
\IfFileExists{footnotehyper.sty}{\usepackage{footnotehyper}}{\usepackage{footnote}}
\makesavenoteenv{longtable}
\usepackage{graphicx}
\makeatletter
\newsavebox\pandoc@box
\newcommand*\pandocbounded[1]{% scales image to fit in text height/width
  \sbox\pandoc@box{#1}%
  \Gscale@div\@tempa{\textheight}{\dimexpr\ht\pandoc@box+\dp\pandoc@box\relax}%
  \Gscale@div\@tempb{\linewidth}{\wd\pandoc@box}%
  \ifdim\@tempb\p@<\@tempa\p@\let\@tempa\@tempb\fi% select the smaller of both
  \ifdim\@tempa\p@<\p@\scalebox{\@tempa}{\usebox\pandoc@box}%
  \else\usebox{\pandoc@box}%
  \fi%
}
\def\fps@figure{htbp}
\makeatother
\usepackage{bookmark}
\IfFileExists{xurl.sty}{\usepackage{xurl}}{} % add URL line breaks if available
\hypersetup{
  pdftitle={Probabilistic Robustness in Medical Image Classification},
  hidelinks,
  pdfcreator={LaTeX via pandoc}}

\title{Probabilistic Robustness in Medical Image Classification}
\author{}
\date{}

\begin{document}
\maketitle

Yi Zhang\textsuperscript{1}, Siddartha Khastgir\textsuperscript{1},
Xingyu Zhao*\textsuperscript{1}

\textsuperscript{1}WMG, University of Warwick, Coventry, United Kingdom

\{yi.zhang.16,
\href{mailto:s.khastgir.1,\%20xingyu.zhao\%7d@warwick.ac.uk}{s.khastgir.1,
xingyu.zhao\}@warwick.ac.uk}

*Corresponding author: Xingyu.Zhao@warwick.ac.uk

\section{Abstract}\label{abstract}

Deep learning (DL) has shown strong performance in medical image
classification, but its trustworthy deployment remains challenging in
safety-critical clinical settings, where prediction errors under
perturbations may lead to severe consequences. Existing studies mainly
focus on adversarial robustness (AR) from a worst-case perspective;
however, such settings may be less representative of real medical
applications. In this work, we investigate probabilistic robustness (PR)
as a more practical measure of model trustworthiness. To this end, we
construct a set of natural corruption settings for medical image
classification and systematically evaluate commonly used DL models on
MedMNIST v2 dataset. Our study provides a statistically grounded
perspective on assessing the trustworthiness of DL models, thereby
supporting their more trustworthy deployment in medical imaging
applications.

\textbf{Keywords:} probabilistic robustness, trustworthy AI, medical
image classification, natural corruptions

\section{Introduction}\label{introduction}

Deep Learning (DL) has achieved remarkable success in medical image
analysis and has been widely applied in automated disease diagnosis
systems~\cite{lo1995artificial,gulshan2016development}. Its strong representation capability has enabled
image-based applications across a wide range of tasks, including the
analysis and processing of MRI~\cite{korolev2017residual}, CT scans, X-ray images~\cite{rajpurkar2017chexnet},
and skin images for cancer diagnosis, lung disease classification, and
brain tumour identification. As a result, DL-based systems are
increasingly being considered to assist clinical decision-making and
reducing the workload of medical professionals. However, concerns remain
regarding the trustworthiness~\cite{zhang2026trustworthy} of DL-based medical diagnosis
systems under potential attacks, since inaccurate diagnoses may lead to
disastrous consequences in safety-critical settings. Therefore,
robustness is a fundamental prerequisite for the widespread deployment
of DL models in medical applications. Accordingly, numerous studies have
investigated robustness in this context.

Regardless of the specific task or model architecture, robustness
generally refers to a model's ability to maintain consistent decisions
under small input perturbations. A small perturbation on an input is
termed an Adversarial Example (AE) if it leads to a different prediction
from the ground-truth label assigned to the original input. Most
existing studies on robustness focus on the question of maximum
prediction loss, asking: \emph{``which AE within a norm ball yields the
highest prediction loss?''}. Evidently, this formulation emphasizes
\emph{extreme scenarios}, where local robustness is assessed based on
the existence of a deterministic AE, thereby reflecting a worst-case
perspective. A more recent and distinct view, however, considers
probabilistic robustness (PR), employing statistical approaches to its
evaluation~\cite{webb2018statistical,zhang2025probabilistic,zhang2024protip}, it addresses the question: \emph{``what is the
likelihood of AEs in the given perturbation norm-ball?''}~\cite{zhang2025adversarial,zhao2025probabilistic,wang2025non}. This probabilistic view is arguably more relevant to real-world
applications than AR, as it provides an overall assessment of a
model\textquotesingle s local robustness, accounting for scenarios where
AEs may exist and acknowledging residual risks that are more realistic
to manage in practice.

To this end, we conduct a comprehensive investigation of the PR of
commonly used DL models for medical diagnosis tasks. First, we introduce
a set of standard natural corruption settings for medical image
classification by adapting perturbation types widely studied in general
computer vision to the medical imaging domain. We then evaluate a series
of representative DL models with strong classification performance on
the widely used medical dataset MedMNIST v2~\cite{yang2023medmnist}, a large-scale
lightweight benchmark for 2D and 3D biomedical image classification.
Rather than proposing a new classification architecture, our goal is to
establish a principled evaluation framework that complements
conventional performance metrics and offers additional evidence for
understanding the trustworthiness of DL models in medical image
classification. In summary, our main contributions including: (1) we
investigate PR in the context of medical image classification and
highlight its practical relevance for safety-critical medical diagnosis
systems under realistic perturbation scenarios; (2) we construct and
standardize a set of natural corruption settings for medical image tasks
by adapting commonly studied perturbations from general computer vision,
thereby enabling a more realistic robustness evaluation protocol for
medical imaging models; (3) we conduct a systematic evaluation of
several commonly used DL models on MedMNIST v2, analysing their PR under
medical-image corruption settings in addition to conventional predictive
accuracy performance.

\section{Preliminaries and Related
Works}\label{preliminaries-and-related-works}

Medical image classification~\cite{lo1995artificial} is one of the most important
applications of DL in medical image analysis and computer-aided
diagnosis (CAD). In this setting, medical images are used as inputs to
DL models, which assign them to predefined diagnostic categories, such
as disease-positive versus disease-negative cases or multiple disease
classes. Fig.1 illustrates a typical example of using CheXNet~\cite{rajpurkar2017chexnet}
for chest X-ray classification. This paradigm has been widely adopted
across different imaging modalities. Typical applications include lung
disease screening from chest X-rays, diabetic retinopathy detection from
fundus images~\cite{gulshan2016development}, and neurological disorder diagnosis from brain
MRI~\cite{korolev2017residual}. Owing to the strong representation capability of DL models,
medical image classification has become a fundamental tool for assisting
disease diagnosis. Despite its strong performance, concerns regarding
trustworthiness remain. A key requirement for their trustworthy
deployment is robustness, namely the ability of a DL model to maintain
accurate predictions under perturbations. Robustness is therefore a
fundamental prerequisite for the widespread deployment of such models in
medical settings.

\begin{figure}
\centering
\includegraphics[width=6.22917in,height=1.46569in]{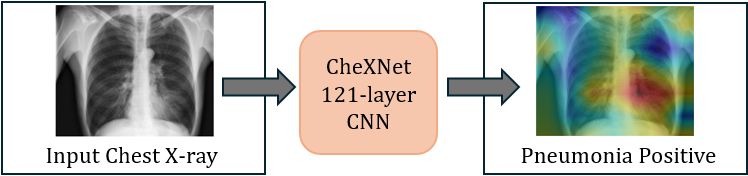}
\caption{Figure 1. An example of CheXNet, a 121-layer CNN, for chest
X-ray classification.}
\end{figure}

\section{Probabilistic Robustness}\label{probabilistic-robustness}

Robustness generally refers to a model's ability to maintain stable
predictions under small input perturbations. In classification task,
robustness is typically studied within a local region around an input
\(x\), defined as an \(L_{p}\)-norm ball
\begin{equation}
    \eta\  = \ \{ x'|\ \| x' - \ x\|\  \leq \ \gamma\}
\end{equation}
where \(\gamma\) denotes the perturbation radius. A perturbed input
\(x' \in \eta\) is regarded as an AE if it causes the model to predict a
label different from the ground-truth label \(y\).

As illustrated in Fig.2(a), AR is commonly studied from a worst-case
perspective. Let \(x \in X \subseteq \mathbb{R}^{d}\)denote an input,
\(y \in Y \subseteq \left\{ 1,2,\ldots,\kappa \right\}\ \)represent the
label, \(D\) be an unknown data distribution over \(X \times Y\), and
\(f_{\theta}:X \rightarrow \mathbb{R}^{\kappa}\) be a DL model
parameterized by \(\theta\). Given a loss function \(L\), AR evaluates
the worst-case perturbation within a norm ball of radius \(\gamma\) by
solving for the AE that maximizes the loss as
\begin{equation}
\delta^{\star} = \arg{\max_{\parallel \delta \parallel \leq \gamma}L(x + \delta,y;\theta)}.
\end{equation}

PR adopts a probabilistic view of robustness by evaluating the overall
local robustness of a model in the presence of AEs, as illustrated in
Fig. 2(b). The formal definition of PR is given as follows.

\begin{figure}
\centering
\includegraphics[width=5.23889in,height=2.39636in]{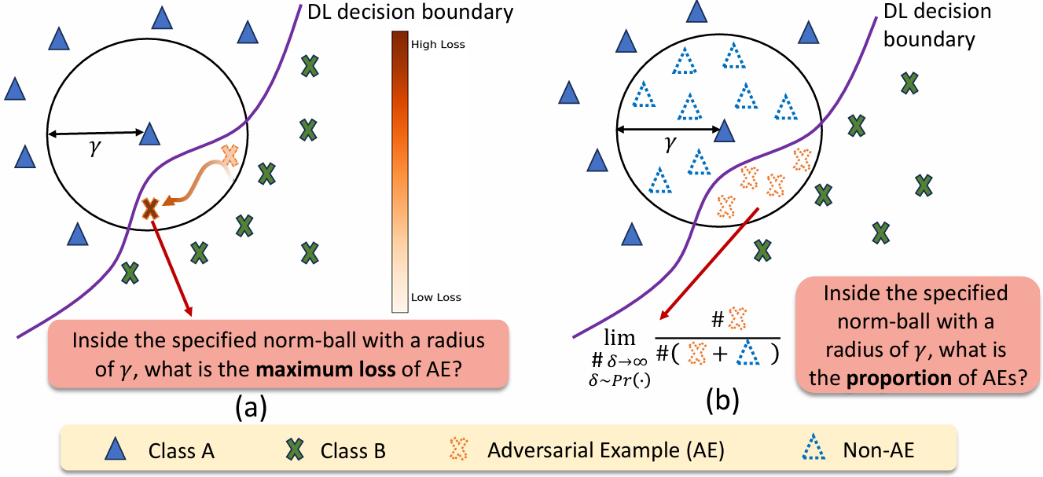}
\caption{Figure 2. Adversarial (a) vs. Probabilistic (b) Robustness.}
\end{figure}

\textbf{Definition 1.} \textbf{(Probabilistic Robustness).} \emph{For a
DL classifier} \(f_{\theta}\) \emph{that takes an input} \(x\) \emph{and
outputs a predicted label, the PR of} \(x\) \emph{within a norm ball of
radius} \(\gamma\) \emph{is defined as}

\begin{equation}
R(x,\ \gamma)\  = \ \mathbb{E}_{\begin{array}{r}
\delta\ \sim\ Pr( \cdot \ |x) \\
||\delta||\  \leq \ \gamma
\end{array}}\ \lbrack\ I_{\{ f_{\theta}(x\  + \ \delta = y)\}}(x\  + \ \delta)\ \rbrack.
\end{equation}

\emph{Here,} \(I_{S}\)\emph{(X) is an indicator function that equals 1
if condition} \(S\) \emph{is true and 0 otherwise.}
\(Pr( \cdot \ |\ x)\) \emph{denotes the local input distribution that
specifies how perturbations} \(\delta\) \emph{are generated; this
corresponds to the ``input model'' used in previous work.}

This definition indicates that PR is the probability that the model
prediction remains unchanged under a random perturbation \(x'\). From a
frequentist perspective, this expected probability can be interpreted as
the limiting relative frequency of perturbations for which the output
label is preserved, over an infinite sequence of independently generated
perturbations. In other words, PR can be viewed as the proportion of
non-AEs in the infinite set of perturbed inputs.

\section{Experiments}\label{experiments}

We evaluate PR on PathMNIST from MedMNIST v2~\cite{yang2023medmnist}, a 9-class
colorectal pathology image classification dataset with 107,180 samples,
split into 89,996/10,004/7,180 for training, validation, and testing,
respectively. Experiments are conducted with two commonly used CNN
backbones, ResNet-18 and ResNet-50. Standard classification performance
is measured by AUC and ACC, following the MedMNIST v2 protocol, while PR
is assessed using the benchmarking protocol defined in PRBench~\cite{zhang2025probabilistic}.

\section{Results \& Discussion}\label{results-discussion}

\textbf{Natural corruptions substantially undermine the trustworthiness
of CNNs.} As shown in Table 1, although the commonly used ResNet-18 and
ResNet-50 achieve high performance on clean images, e.g., ResNet-18
attains 90.68\% ACC and 98.46\% AUC, their PR performance under the six
evaluated natural corruptions is substantially lower than their
clean-image accuracy. In particular, the PR drops to only 62.84\% and
76.83\% under brightness variations. Since such perturbations commonly
arise in real-world scenarios, for example due to camera defocus or
equipment degradation, image quality is often inevitably affected in
practice, which may lead to less trustworthy predictions.

\textbf{PR is corruption dependent.} Both models remain relatively
robust under Pixelate, Stain, and Saturate perturbations, but are much
more vulnerable to Motion blur, with Defocus and brightness variations
also causing substantial degradation, indicating that blur-related
degradations are substantially more challenging than colour or
appearance variations for this task. Moreover, although ResNet-50
achieves a higher ACC/AUC, it does not consistently show better PR,
suggesting that stronger clean performance does not necessarily imply
stronger robustness under realistic corruptions.

\begin{table}[htbp]
\centering
\caption{Clean-image performance (ACC/AUC) and PR performance under six natural corruptions on PathMNIST.}
\label{tab:pathmnist-pr}
\begin{tabular}{lcccccccc}
\toprule
Model & ACC & AUC & Defocus & Motion & Stain & Saturate & Pixelate & Bright +/- \\
\midrule
ResNet-18 & 90.68 & 98.46 & 66.03 & 40.20 & 84.07 & 77.03 & 80.97 & 62.84 / 76.83 \\
ResNet-50 & 92.05 & 99.39 & 72.32 & 40.74 & 83.41 & 76.89 & 80.19 & 67.91 / 77.03 \\
\bottomrule
\end{tabular}
\end{table}

\section{Conclusion}\label{conclusion}

In this work, we investigated probabilistic robustness for medical image
classification and highlighted its practical relevance for trustworthy
medical diagnosis systems. We constructed a natural corruption setting
for medical image tasks, systematically evaluated commonly used deep
learning models on MedMNIST v2. Our study shows that PR offers a more
practical perspective than worst-case robustness for assessing model
trustworthiness under realistic perturbations and provides a
statistically grounded reference for the safer deployment of deep
learning systems in medical imaging.

\bibliographystyle{unsrt}
\bibliography{ref}

\end{document}